\documentclass{article}

\usepackage{amsmath}
\usepackage{amsfonts}
\usepackage{microtype}
\usepackage{graphicx}
\usepackage{booktabs} 
\usepackage{mathtools}
\usepackage{todonotes}
\usepackage{chngcntr}
\usepackage{subfig}

\DeclareMathOperator*{\argmin}{arg\,min}
\newcommand{\defeq}{\vcentcolon=}
\usepackage{hyperref}

\usepackage[accepted]{icml2019}

\icmltitlerunning{Towards Principled Uncertainty Estimation for Deep Neural Networks}

\begin{document}

\twocolumn[
\icmltitle{Towards Principled Uncertainty Estimation for Deep Neural Networks}



\icmlsetsymbol{equal}{*}

\begin{icmlauthorlist}
\icmlauthor{Richard Harang}{sophos}
\icmlauthor{Ethan M. Rudd}{sophos}
\end{icmlauthorlist}

\icmlaffiliation{sophos}{Sophos PLC, Reston Virginia, USA}

\icmlcorrespondingauthor{Richard Harang}{first.last@sophos.com}
\icmlcorrespondingauthor{Ethan Rudd}{first.last@sophos.com}

\icmlkeywords{Machine Learning, ICML}

\vskip 0.3in
]



\printAffiliationsAndNotice{}  

\begin{abstract}
When the cost of misclassifying a sample is high, it is useful to have an accurate estimate of uncertainty in the prediction for that sample. There are also multiple types of uncertainty which are best estimated in different ways, for example, uncertainty that is intrinsic to the training set may be well-handled by a Bayesian approach, while uncertainty introduced by shifts between training and query distributions may be better-addressed by density/support estimation. In this paper, we examine three types of uncertainty: model capacity uncertainty, intrinsic data uncertainty, and open set uncertainty, and review techniques that have been derived to address each one. We then introduce a unified hierarchical model, which combines methods from Bayesian inference, invertible latent density inference, and discriminative classification in a single end-to-end deep neural network topology to yield efficient per-sample uncertainty estimation in a detection context. This approach addresses all three uncertainty types and can readily accommodate prior/base rates for binary detection. We then discuss how to extend this model to a more generic multiclass recognition context.  
\vspace{-1em}
\end{abstract}

\section{Introduction}

In practical applications of machine learning, knowing the uncertainty of a prediction can be almost as important as knowing the most likely prediction. For binary classification responses (or calibration thereon) given in a 0-1 range, the distance from one extreme or the other is often taken as a proxy for the certainty (or uncertainty) of the classification. While for a binary cross entropy loss under certain conditions this estimate of uncertainty is correct -- at least in the asymptotic sense that it attains the posterior conditional probability of the label being in the `positive' class --  the general approach of using the output score of a classifier does not typically yield a faithful estimate of uncertainty in the above sense, and does not suggest any degree of uncertainty about the obtained point estimate. 

Furthermore, in the finite-data case, and especially with expressive modern classifiers that apply nonlinear transformations, partitions, or both to the input space, the score itself is subject to a significant degree of uncertainty that is frequently difficult to characterize precisely. Thus, even if we accept the score as a proxy for uncertainty, we may be uncertain about how accurate this measurement of uncertainty is. 

In simpler classifiers, with low-dimensional input spaces, direct estimation of uncertainty can be performed by examining the support of a test point within the training data, but for high-dimensional inputs, the curse of dimensionality can make it difficult and expensive to make an accurate estimate of the support. Even when this difficulty can be overcome, the complex relationships between these inputs 
means that areas of high or low support in the input space may not be so well (or poorly) supported within the transformed space within which the classifier is effectively making its prediction. 

Several methods have been proposed to estimate uncertainty in deep neural networks, including variational methods, Bayesian modeling of stochastic processes, and multi-half space classifiers. While many of these approaches have merit, many are also cumbersome, scaling is questionable, and they address different types of uncertainty with different underlying causes. In this paper, we examine three different types of uncertainty and their underlying causes and seek a unified end-to-end model which addresses them all and that works well at scale. 

We examine the uncertainty estimation problem with a Bayesian perspective in mind and show, for a binary detection problem, how combining deep neural networks as approximate density models with a Bayesian-inspired model can lead to uncertainty estimates for models that are robust, consistent (in a particular empirical sense that we outline below), require comparatively little additional computation to obtain, and can in most cases be directly converted into a maximum a posteriori estimate `score' for the network. We then discuss steps forward to extend this model to generic multiclass uncertainty estimation.

\begin{figure*}[h!]
\centering
\includegraphics[width=.32\textwidth]{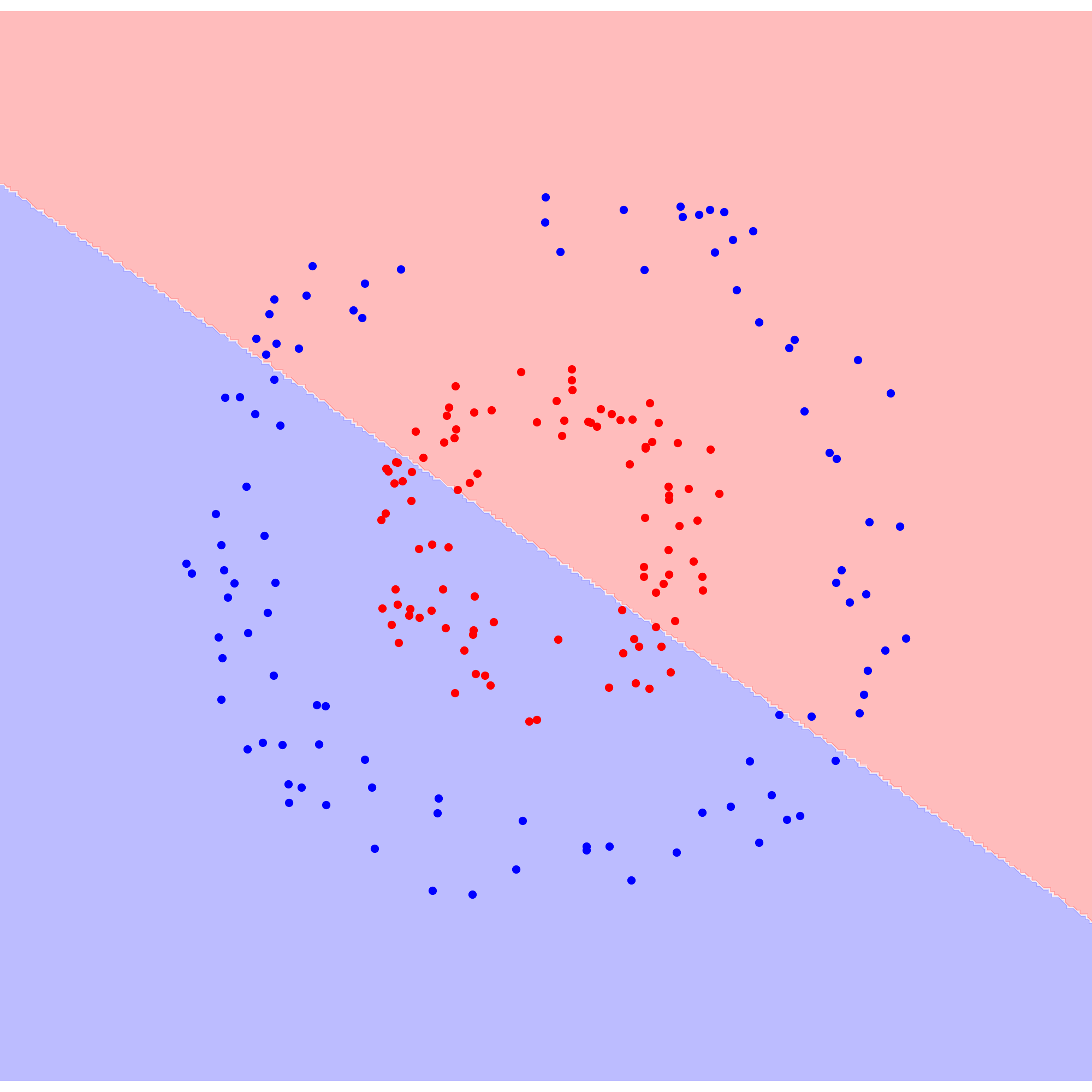}
\includegraphics[width=.32\textwidth]{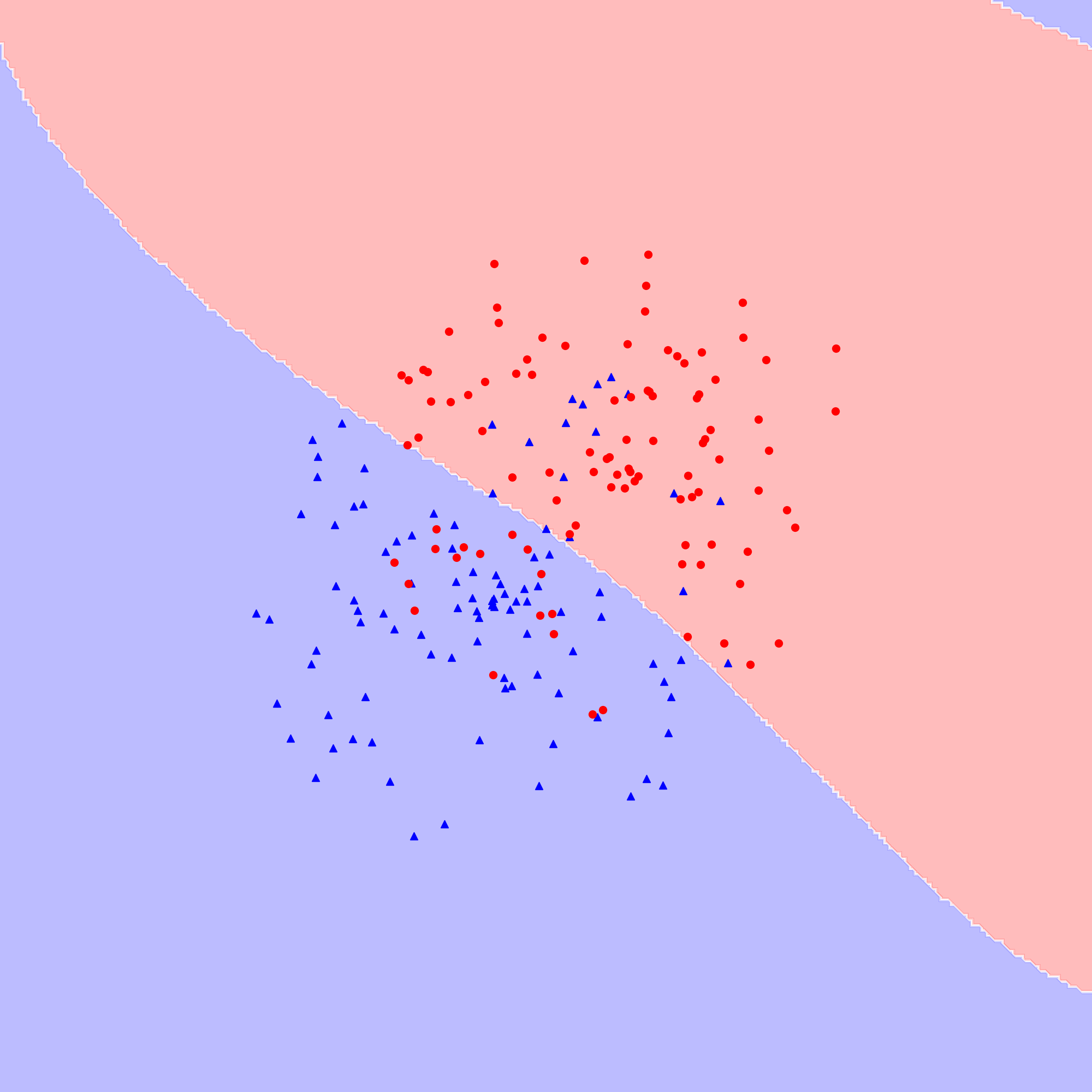}
\includegraphics[width=.32\textwidth]{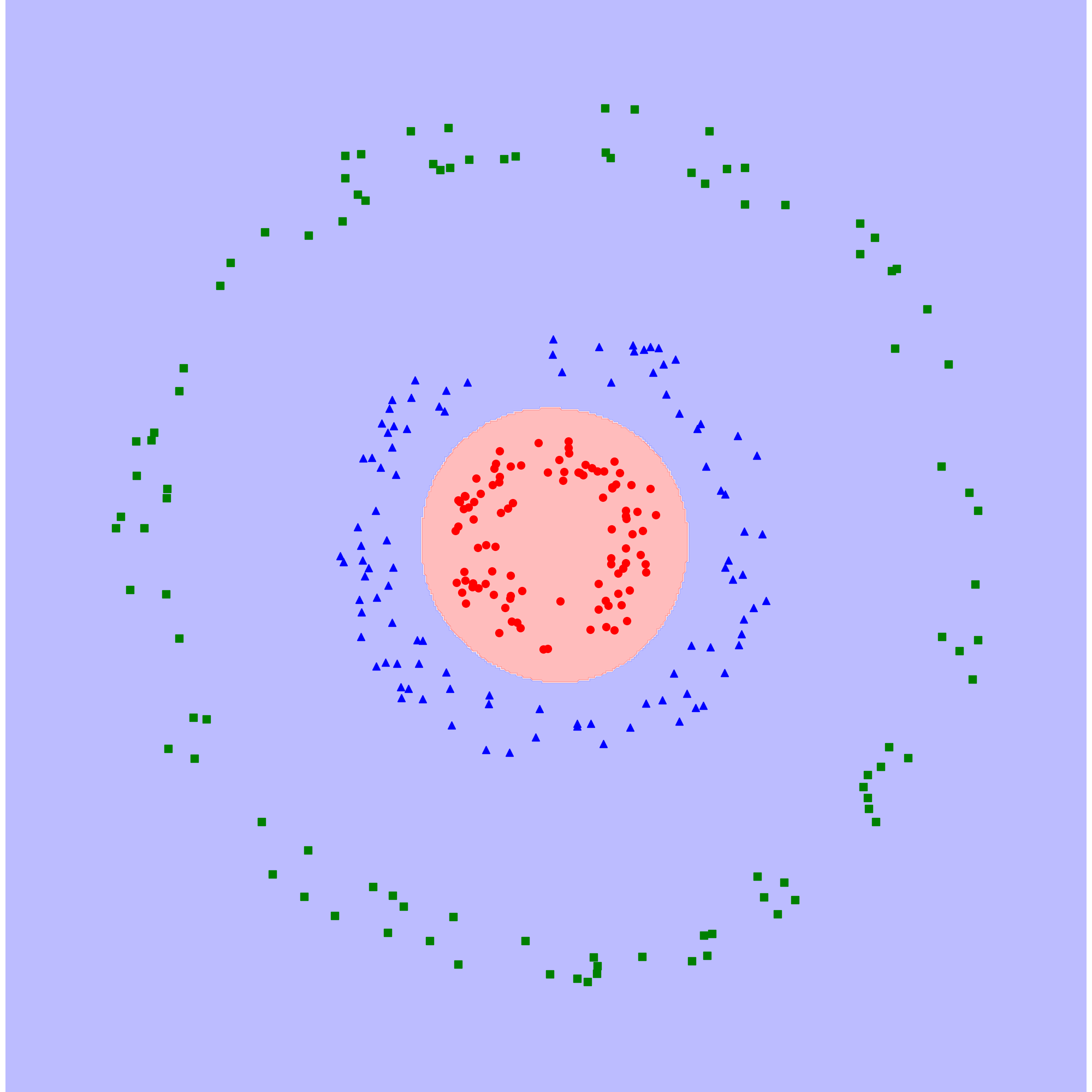}
\caption{In this paper we address these three types of uncertainty simultaneously. {\bf Left: Model Capacity Uncertainty.} The red and blue points are cleanly separable, but the classifier lacks sufficient capacity to separate them along a nonlinear boundary. {\bf Middle: Intrinsic Uncertainty.} A nonlinear classifier of best fit is able to separate the classes within the limit of the data, however the feature space is insufficiently expressive to obtain a better classifier with the given features. {\bf Right: Open Set Uncertainty.} During training on blue and red classes a classifier is fit which  perfectly separates the data, but what if points from a novel green class appear during deployment (these could also correspond to concept-drifted points from the red class)? If the classifier does not bound its decision by the support of the training set, they will be mis-ascribed to the blue class, typically with extremely high confidence.}
\label{fig:uncertainty_types}
\end{figure*}

\vspace{-1em}
\section{Background: Types of Uncertainty}

Model uncertainty can be categorized into three types, which are illustrated in Fig. \ref{fig:uncertainty_types}: 

\begin{enumerate}
    \item {\bf Model capacity uncertainty} is a property of the model and is introduced if the model has too little or too much capacity to fit the data accurately even when the data is intrinsically separable by a Bayes optimal classifier.
    
    \item {\bf Intrinsic uncertainty} with respect to the data. This is a property of the underlying data distribution such that the data is ill-separated even by a Bayes optimal classifier. When intrinsic uncertainty is high, any feature space transformation or classifier which cleanly separates the training data is by definition over-fitting.
    
    \item {\bf Open Set uncertainty} due to distributional discrepancies between training and query data. When the data distribution shifts between training and query time and new modes/classes appear, open set uncertainty is high. This violates a  simplifying assumption common to most classification problems that train and test data are drawn independently from identical distributions.
\end{enumerate}

Model capacity uncertainty is commonly addressed by choosing a model with abundant capacity and then regularizing to prevent over-fitting \cite{Bishop:2006:PRM:1162264}. In neural networks, regularization amounts constraining the range of the weights in the optimization algorithm. In effect, regularization aims to jointly minimize fitting error and model complexity to arrive at the simplest model which accurately explains the data. For neural networks, popular techniques include $l_p$-norm  regularization \cite{Bishop:2006:PRM:1162264}, dropout \cite{srivastava2014dropout}, batch normalization \cite{ioffe2015batch}, and layer normalization \cite{ba2016layer} to name a few.

Several approaches have been applied to intrinsic uncertainty estimation. Score calibration methods \cite{platt1999probabilistic,niculescu2005predicting,scheirer2010robust} aim to return a posterior prediction $p(label|score)$, but in doing so, they make strong distributional assumptions on scores, the sampling of data, and the goodness of fit of the model, ignoring much information present in the model representation and the input feature space. Jiang et al. formulated a \textit{trust score} in feature space, using ratios of distances between nearest neighbors of different classes as a proxy for score confidence \cite{jiang2018trust}. However, their reliance on nearest-neighbor approaches incurs a computational cost in high dimensions, and produces `trust scores' for a classification, suggesting a probability of \emph{error}, rather than a measure of \emph{uncertainty}.  

Several Bayesian approaches have been developed which aim to engineer intrinsic uncertainty estimates directly into models. Gaussian processes model uncertainty over functions generated by a stochastic process, by assuming an $N$-dimensional Gaussian joint distribution on function values over $N$ ``context" points thought to have been generated by the process. This serves a prior for additional observations from which a posterior can be derived \cite{williams1996gaussian}. More recently, Conditional Neural Processes \cite{garnelo2018conditional} and Neural Processes \cite{garnelo2018neural} instead model the function generating stochastic process via a neural network embedding and variational inference. Gal et al. \cite{gal2015dropout} derive a relationship between Bayesian inference and dropout-regularized neural networks, in which dropout can be viewed as a Bayesian approximation which is easy to sample from to obtain a posterior distribution. In practice, however, their argument is limit-based and the posterior estimates require multiple stochastic forward passes through the network, increasing computational costs.

Several approaches aim to address open set uncertainty by either 1) incorporating training set support into the model's optimization process  \cite{scheirer2013toward,rudd2018extreme,kardan2016fitted} or 2) estimating training set support post-hoc and exercising a rejection option as necessary \cite{scheirer2014probability,jiang2018trust}. 
Scheirer et al. were the first to formalize the \textit{open set problem}, and addressed it by fitting slabs of hyperplanes with a linear SVM jointly optimized to separate classes in the training set and avoid ascribing labels to unsupported hypothesis space \cite{scheirer2013toward}. In later work, they extended the open set paradigm to provide probabilistic outputs for nonlinear problems \cite{scheirer2014probability}. Rudd et al. \cite{rudd2018extreme} and Bendale and Boult \cite{bendale2016towards} applied post-hoc open set probability estimators in a deep feature space, but these models are not jointly optimized with the network. By contrast, Kardan and Stanley fit an end-to-end overcomplete network, which leverages an intersection of several hyperplanes at the output layer of the network to limit the labeling of unsupported space \cite{kardan2016fitted}, but it is not immediately obvious that their approach scales to high-dimensional problems and it does not provably bound open space risk \cite{scheirer2013toward}. 

As Rudd et al. discuss in \cite{rudd2017survey}, density estimation approaches in the feature space can also be used as open set uncertainty estimators. However, for high-dimensional input spaces, this demands some kind of sampling or approximate inference. Monte Carlo methods aim estimate density by sampling, but tend to over-sample certain modes while missing others if step size and number of steps are not properly tuned. Moreover, the number of steps required for good convergence is often not immediately obvious. Approximate inference methods have their own issues, tending to ignore details in improbable parts of the approximated distribution. To address these limitations, in this paper, we leverage a bijective flow-based density-estimation method that uses a special case of a Jacobian change of variables technique to perform distributional modeling in a perfectly invertible latent space \cite{dinh2016density}.
\vspace{-1em}
\section{Theoretical Foundations}
To arrive at a theoretically grounded uncertainty estimator which addresses multiple uncertainty types, we combine several techniques into one end-to-end model. In Sec. \ref{sec:glm}, we review how neural networks can be used to estimate generic parameters of a noise model beyond just the maximum likelihood estimate of the label of interest, including uncertainty parameters. We then discuss how to combine base rates and estimates for parameters of a conjugate prior model to model intrinsic uncertainty in a more theoretically defensible manner in  Sec. \ref{sec:conjugate_prior}. However, this model still requires an estimation of per-class counts, which must be robust to realistic open set assumptions. In section \ref{sec:flow} we first review flow-based models for density estimation, and then introduce a variation on flow-based models which fits a separate density model for each class and thus allows an approximation of density for a given class.  These density estimates can be used in a number of ways: we focus on the use of the density estimates to estimate the likelihood of a given sample being within the set (addressing open-set uncertainty) and -- via count approximation -- to derive an estimate for intrinsic uncertainty.

\subsection{Neural Networks as Estimators for Parametric Distributions}
\label{sec:glm}

Neural networks are often employed in practice to estimate labels for samples, but more  fundamentally, they provide point estimates of distributional parameters under a presumed model on label noise (e.g., Gaussian for regression using mean squared error loss, Bernoulli for binary detection problems, multinomial for multi-class recognition). This link is made explicit in the case of Generalized Linear Models (GLMs) \cite{mccullagh1989generalized} in which linear functions are passed through a potentially nonlinear transform to model the parameters of a particular distribution associated with the observed data.  Often, the output of the model and associated loss function imply that the mean is predicted while other parameters are ignored. In this paper, however, we are interested in deep neural networks as generic estimators. 

Consider a generic parameter $\varsigma$ that we are trying to estimate using a model with trainable parameters (weights) $\theta$, e.g., the mean value of a prediction for a sample $x$. For a linear model where $\theta$ is a vector, $x_0 \defeq 1$, and $\theta_0$ is the bias term, $\varsigma=g(\theta^T x)$, where $g(\cdot)$ is a non-parametric \textit{link function} chosen to modify the output of $\theta^T x$ to appropriate ranges for $\varsigma$.\footnote{For continuous parameters with arbitrary range, e.g., the mean of a Gaussian, $g(\cdot)$ is an identity function, while for parameters that must be non-negative, e.g., the variance of a Gaussian distribution, an exponential/log link is typically used: $\theta^T x = log(\varsigma)$; $\varsigma = exp(\theta^T x)$.}


As an example, consider linear regression under a Gaussian noise model: it is common to estimate the mean, but we can also estimate the variance. Consider a data set consisting of samples $X$ with corresponding labels $Y$. Given $i$th sample $x_i \in X$  with $i$th label $y_i \in Y$, we can express the per-sample mean estimate $\mu_i | x_i$ and per-sample variance estimate ${\sigma^2}_i | x_i$ in terms of inner products on trainable parameter vectors and sample feature vectors:

\begin{align}
    \mu_i &= {\theta_1}^T x_i\label{eq:mu}\\ 
    {\sigma_i}^2 &= exp({\theta_2}^Tx_i).\label{eq:sigma}
\end{align}

We then select $\theta_1$ and $\theta_2$ to maximize the likelihood of the parameters given the data. Recalling the Gaussian distribution and assuming independence yields the following likelihood over the dataset:

\begin{equation}
\label{eq:gll}
    \mathcal{L}(\theta_1,\theta_2;X,Y) = \prod_{i=1}^{|X|} \frac{1}{\sqrt{2\pi{\sigma_i}^2}}exp-\left(\frac{(y_i-\mu_i)^2}{2{\sigma_i}^2}\right).
\end{equation}

Maximizing this function is equivalent to minimizing the negative log likelihood loss function: 
\begin{equation}
\label{eq:gaussian_nll}
-log(\mathcal{L}(\theta_1,\theta_2;X,Y)) = \sum_{i=1}^{|X|} log(\sqrt{2\pi{\sigma_i}^2}) + \frac{(y_i-\mu_i)^2}{2{\sigma_i}^2}.
\end{equation}

Thus, we seek $\argmin_{\theta_1,\theta_2} -log(\mathcal{L}(\theta_1,\theta_2;X,Y))$. Carrying out that minimization allows us to estimate both mean \textit{and} variance.  We can trivially extend these estimators for per-sample mean and variance to neural networks by substituting the output of the  penultimate hidden layer for $x_i$ in Equations \ref{eq:mu}-\ref{eq:gaussian_nll}. In practice, $\theta_1$ and $\theta_2$ are jointly optimized with the rest of the network parameters via backpropagation. An example result is shown in Fig. \ref{fig:gaussian_glm} of the Appendix.
While this approach to variance estimation is not statistically well-grounded, and is presented in a regression context, in this paper we ask, can we derive a similar lightweight technique that uses neural networks as generic parameter estimators, extends to the classification context, and is more theoretically principled?

One might imagine, for instance, applying a similar technique to a binary classification task by fitting the parameters of a Beta distribution in order to obtain a density estimate for the probability of inclusion in one class or the other. Such a technique could have the added advantage of being able to incorporate priors/base rates and provide a more statistically principled treatment of uncertainty estimation.

\subsection{Conjugate Prior Bayesian Models}
\label{sec:conjugate_prior}
Given a prior distribution $p(\varsigma)$ on some generic parameter $\varsigma$, observation $y$ with evidence $p(y)$, and likelihood $p(y|\varsigma)$, Bayes rule states that the posterior distribution of $\varsigma|y$ is given by 
$p(\varsigma|y) = \frac{p(y|\varsigma)p(\varsigma)}{p(y)}$.

It is often mathematically convenient to select a combination of prior and generative model for $y$ such that the posterior will have the same distributional form for a given likelihood. Such a prior is referred to as a \textit{conjugate prior}. 

For binary classification, we assume that labels are distributed under a Bernoulli noise model per sample:

\begin{equation}
    p(y_i|\mu_i) = {\mu_i}^{y_i}(1-\mu_i)^{1-y_i},
\end{equation}
\noindent 
which takes a Beta distribution, with parameters $a$ and $b$ as a conjugate prior for $\mu_i$:
\begin{equation}
    p(\mu_i) = \beta (\mu_i|a,b) = \frac{\Gamma(a+b)}{\Gamma(a)\Gamma(b)} \mu_i^{a-1}(1-\mu_i)^{b-1},
\end{equation}
\noindent
where $\Gamma(\cdot)$ is a Gamma function. The posterior distribution will then be defined by $p(\mu_i|y_i) = \beta(a',b')$ with the values of $a'$ and $b'$ dependent on $y_i$.  Note that from such a \emph{distributional} estimate we may not only derive point estimates such as the mean or median of the posterior distribution over $\mu_i$, but also estimates related to \emph{uncertainty} such as the width of a 95\% credible set. 
Note also that $a$ and $b$ and $a'$ and $b'$ can be interpreted as counts of positive or negative samples, and it is thus easy to enforced a prior on the Beta distribution by modifying these respective parameter values and updating our posterior with new evidence respectively. For multiclass classification, we may extend the above by use of a Dirichlet prior on a multinomial noise model with similar results, however we omit detailed examination of this aspect due to space. 

The application of this model to uncertainty estimation has some difficulties, however.  Consider the case in which we have an (unknown) binary response $y_i\in \{0,1\}$ with predictor $x_i \in \mathcal{R}^N$ and wish to estimate the \emph{distribution} of $\hat{\mu_i}=P(y_i=1|x_i)$.  To directly apply a Beta-Binomial model as above, we require some estimate of `counts' of samples of class $y_i=1$ and $y_i=0$ within some region about $x_i$.  In the case of high-dimensional predictors, however, counts will be sparse in any region, making nearest-neighbor approaches infeasible and difficult to scale.  In order to find a useful proxy for counts under such conditions, we turn to density estimation.


\subsection{Class-Conditional Flow-Based Models}
\label{sec:flow}

In practice, determining probability densities in a high-dimensional input space  directly is a challenging task. One way of dealing with this is to transform the input space into a latent space  where samples approximately exhibit a distribution of choice, perform density estimates, and then transform back to the input space. While this task can be accomplished using certain flavors of autoencoders (such as an adversarial autoencoder \cite{makhzani2015adversarial}), dimensionality-reducing lossy transformations between the input space and the latent space are potentially problematic for intrinsic uncertainty estimation because effects of outliers can be attenuated or completely removed. This may significantly impact uncertainty estimates, particularly in areas of high intrinsic uncertainty.

In this work, we turn to flow-based models --  fully-invertible generative models which employ bijective transformations that can be used to evaluate densities on a sample-wise basis. Several types of flow-based models have been derived, including \cite{rezende2015variational,dinh2014nice,dinh2016density,kingma2018glow}. Each of these aim to simplify the change-of-variables technique by constraining the functional form of the bijective transformation from the input space $X$ to the transformed space $Z$. Let us recall that the change of variables technique can be used to evaluate densities via point-wise transformations scaled by the determinant of the Jacobian of the transforming function with respect to the original space as follows:  

\begin{equation}
    p_X(x) = p_Z(z) \frac{1}{\left| det\left(\frac{\partial f(x)}{\partial x^T} \right) \right|}.
    \label{eq:change_of_variables}
\end{equation}

\begin{figure}[h!]
\centering
\includegraphics[width=\linewidth]{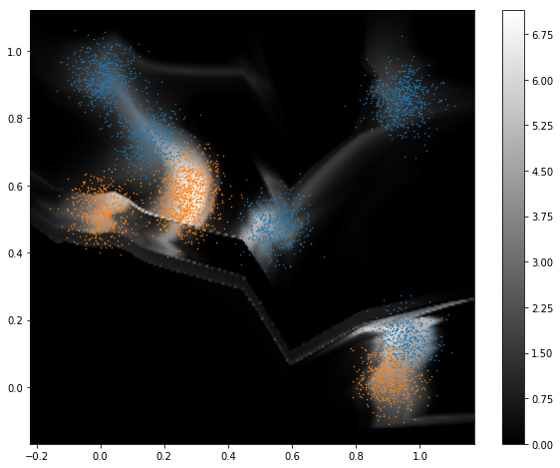}\\
\includegraphics[width=0.48\linewidth]{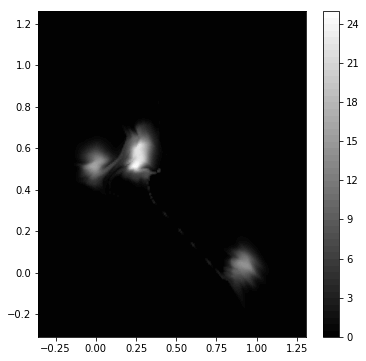}
\includegraphics[width=0.48\linewidth]{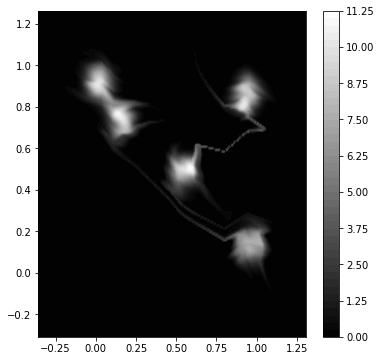}
\caption{Top: Total data density for a toy data set estimated by training a real-NVP stack using a Gaussian negative log likelihood loss. Bottom: Using per-class NVP stacks atop a common base topology, we can obtain class-conditional density estimates. Note that jointly optimizing class conditional estimators and incorporating a standard discriminative loss atop the base drives class separation in the hidden layer space and empirically improves the compactness of the class-conditional distributions.}
\label{fig:rnvp}
\end{figure}

In practice, this requires a transformation with a stable and computationally tractable Jacobian determinant which also results in an easy-to-parameterize latent space distribution (e.g., Gaussian). While multiple authors present techniques for doing so \cite{rezende2015variational,dinh2014nice,dinh2016density,kingma2018glow}, in this work, we utilize real non-volumetric preserving (RealNVP) transformations from \cite{dinh2016density}, in the form of \textit{coupling layers}.

These transformations are bijective element-wise translations by a function $t(\cdot)$ and exponentiated scalings by a function $s(\cdot)$ on chunks of permuted elements of the input vector with respect to each other. Specifically, given input vector $x \in \mathbb{R}^D$, $x_{1:d}$ the first $d$ elements, and $x_{d+1:D}$ the remaining elements with $d < D$, the Real-NVP transformation ($y$) is defined as:

\begin{align}
    y_{1:d}&=x_{1:d}\\
    y_{d+1:D}&=x_{d+1:D} \odot exp(s(x_{1:d})) + t(x_{1:d}).
    \label{eq:coupling}
\end{align}

The resultant Jacobian of a Real-NVP transformation is a diagonal matrix whose determinant can be evaluated as a product over diagonals elements; the computation is unaffected by the complexity of either $s$ or $t$. Thus, $s$ and $t$ can be neural networks of arbitrary capacity themselves, yielding extremely flexible Real-NVP transformations with tractable determinants. Moreover, these transformations are trivially and perfectly inverted and composed, so stacks of coupling layers can be employed for additional capacity. Note also that when stacking coupling layers, a separate permutation on input elements can (and should) be defined at each layer in order to ensure that each element of the data has opportunity to affect each other element.

Taking the negative log likelihood of Eq. \ref{eq:change_of_variables} yields:
\begin{equation}
    -\log(p_X(x)) = - \log(p_Z(z)) + \log\left| det\left(\frac{\partial f(x)}{\partial x^T} \right) \right|.
\end{equation}

Assuming a unit isotropic Gaussian distribution on $p_Z(z)$, i.e., by setting $\mathbf{\mu}:= \mathbf{0}$ and $\Sigma:= \mathbf{I}$ we can train our coupling stack to serve as a density estimator via standard backpropagation.

\begin{figure*}[!t]
\centering
\subfloat[FFNN]{\includegraphics[width=\linewidth]{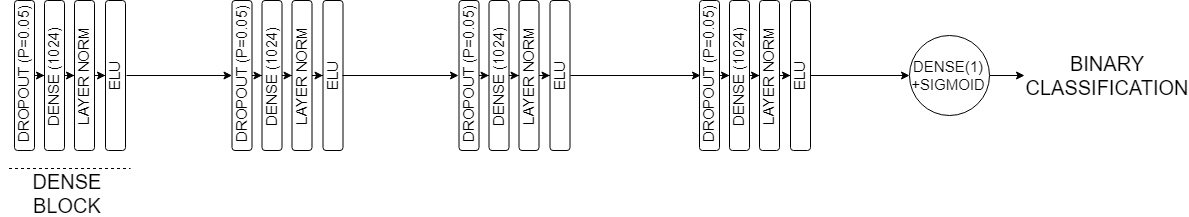}\label{fig:ffnn_topology}}\\
\subfloat[CCCP-DE]{\includegraphics[width=\linewidth]{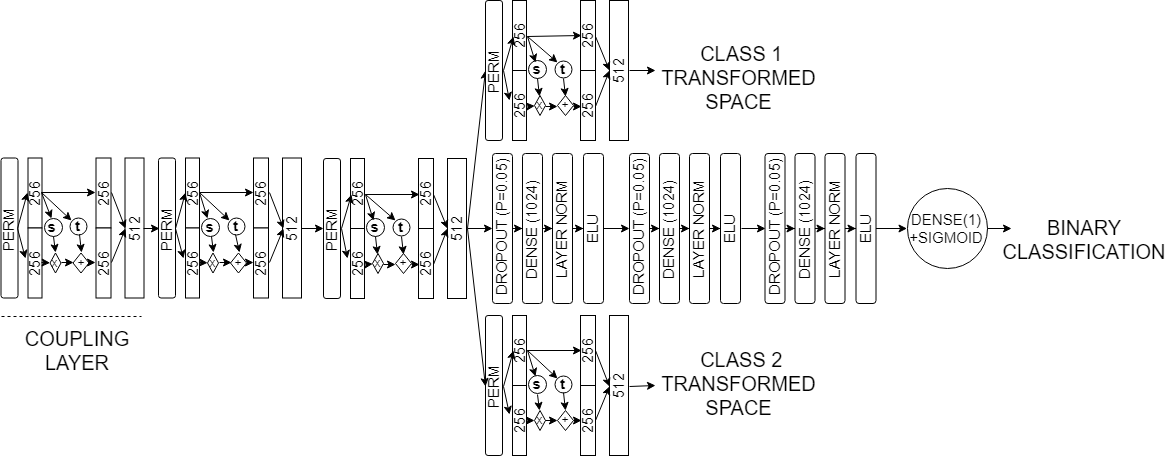}\label{fig:cccp-de_topology}}
\caption{Model topology diagrams. The feed forward neural network  (FFNN)  shown in \protect\subref{fig:ffnn_topology} consists of four dense blocks of dropout, a fully connected layer, layer normalization and an exponential linear unit activation followed by a final dense layer and sigmoid output. Our novel class-conditional conjugate prior density estimator (CCCP-DE), shown in \protect\subref{fig:cccp-de_topology} uses a base stack of coupling layers, followed by one coupling layer per class and a stack of three dense blocks followed by a final dense layer and sigmoid output to guide density estimates to guide the base coupling stack to learn a discriminative subspace. The transformed spaces for each class are fully invertible and are used to assess densities in the input space. These density estimates in turn (or a Monte-Carlo integration thereof) can be employed as proxies for counts in a beta-binomial model.}
\label{fig:net_topologies}
\end{figure*}

While flow-based methods are often used in the context of either generative or unsupervised modeling, work performed concurrently with ours \cite{nalisnick2018deep} has revealed that in many cases, fitting a single density to the entire input data space does not produce high quality density estimates. We vary the usual approach by introducing \textit{class-conditional} heads, i.e.,  placing one stack of coupling layers per-class to estimate class density atop a common coupling layer base topology.  While this increases the number of parameters in the network, we observed empirically in early work that applying flow-based models to inherently multi-modal data led to areas of the feature space ``connecting'' the modes which contained no density receiving significant density, due to the continuous nature of the flow-based transformation of the input space (see, for example, Fig. \ref{fig:rnvp}, and compare the presence of `rabbit trails' in top and bottom plots).  This issue is naturally exacerbated in high-dimensional spaces, and motivated our use of class-conditional heads in an attempt to mitigate this problem.

\begin{figure*}[t!]
\centering
\includegraphics[width=0.49\linewidth,height=0.32\textheight,keepaspectratio]{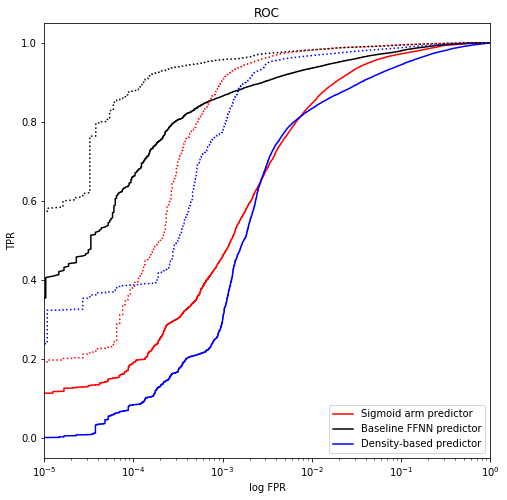}
\includegraphics[width=0.49\linewidth,height=0.32\textheight,keepaspectratio]{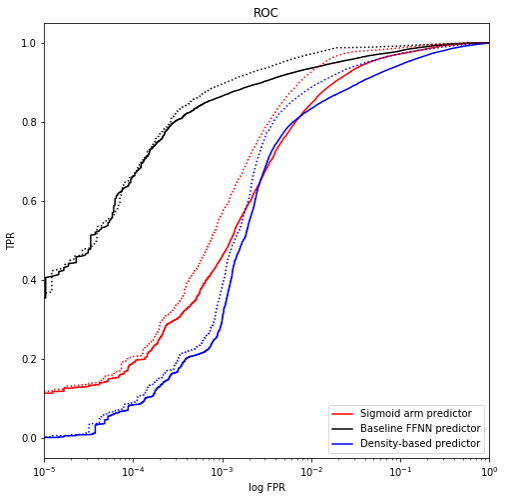}
\caption{
{\bf Left}: Performance before (solid lines) and after (dotted lines) filtering out `uncertain' test samples based on the range of the 95\% credible set obtained via the CCCP-DE. {\bf Right}: Performance before (solid lines) and after (dotted lines) filtering out the same number of samples using the baseline model's score as a proxy for uncertainty. Solid lines represent the ROCs from the predictions of our fully-connected baseline model (black), the sigmoid arm of our principled uncertainty estimator (red), and a prediction from class-densities based on ratio test (blue). Particularly at low false positive rates, gains achieved via the CCCP-DE model are substantially more noticeable than those achieved using scores as a proxy for uncertainty.
}
\label{fig:branched_roc}
\end{figure*}

In addition to mitigating the risk of assigning high density to out-of-class regions of the input space, generating independent densities per class allows us to estimate relative likelihoods for a given test point.  Using a technique such as Monte Carlo integration, we can estimate the number of samples we might obtain under further sampling in a given neighborhood, and thus use the methods presented in the previous sections to obtain intrinsic uncertainty estimates.  In practice, we find that for most problems simply computing point-wise estimates of likelihood, optionally with rescaling based on the class frequencies, leads to acceptable results, though proxy methods which push the densities to be more similar to probabilities could also be employed \cite{theis2015note}. We describe the construction of our model in more detail in the following section.

\section{Experiments}
\label{sec:experiments}

In this section, we evaluate our principled uncertainty estimation technique in a binary classification context. To estimate class-conditional densities, we employ  multi-headed neural networks over a shared base topology, with one head trained to minimize Gaussian negative log likelihood loss for each respective class. In all cases, the shared base topology consisted of NVP coupling stacks. To simplify the task of the class-conditional heads, we added an additional head with a sigmoid output trained on a conventional cross-entropy loss.

We consider the malware detection task, a task where certainty/uncertainty of the classification is practically  relevant:  In deployment scenarios, where static binary detectors perform bulk scans, with limited budget for dynamic analysis in a sandbox, suspicious samples can be flagged for sandbox runs according to their uncertainty. Likewise, in an R\&D or forensics context, high uncertainty samples can be flagged for manual inspection or relabeling. 

Using a vendor aggregation service, we collected a data set of portable executable (PE) files with unique SHA256 hashes, containing 3 million files from January and an additional 3 million files from February of 2018.  January files were used as a training set and February files as a test set. As features, we extracted byte entropy and hashed imports from the import address table (IAT) of each executable (cf. \cite{saxe2015deep}). Our byte entropy histogram features consisted of 256 bins, with 16 uniformly split across byte and entropy axes. Our net input feature vector size was 512. Models were trained for 10 epochs each using the Adam optimizer and a minibatch size of 512.

We evaluated two different types of models: a baseline model and our attempt at a principled uncertainty estimator which uses class-conditional outputs as a proxy for counts in a Beta-Binomial model; for brevity we refer to it henceforth as a Class-conditional/conjugate prior density estimator (CCCP-DE). 

For our baseline model (Fig. \ref{fig:ffnn_topology}), we used a fully connected feed-forward neural network (FFNN) of similar topology to \cite{saxe2015deep,rudd2018meade}, consisting of four \textit{dense blocks} of dropout with a dropout ratio 0.05, followed by a linear dense layer (1024-dimensional), layer normalization, and an exponential linear unit (ELU) activation, with a sigmoid output and a binary cross entropy loss function.  This model was trained for 10 epochs using the Adam optimizer as a standard classification model.

For our CCCP-DE model (Fig. \ref{fig:cccp-de_topology}), we used a stack of three NVP coupling layers, as a base, followed by one NVP coupling stack with Gaussian negative log likelihood loss for malicious and benign samples respectively. Finally, we stacked three dense blocks of the same form as our fully-connected model followed by a sigmoid output layer atop the base NVP stack to form a third auxiliary loss. Each coupling layer consists of a 512-dimensional input, and a permutation on that input (stored to maintain invertibility); the input is split in half, with one half maintained and passed through translation $t$ and scaling $s$ functions Eq. \ref{eq:coupling}. The other half is point-wise multiplied with the exponentiation of the scaling on the first half and added to the translation on the first half. For $s$ and $t$ we used  fully-connected networks with three dense layers, the first two followed by leaky ReLU activations, the output followed by a $tanh$ activation, with 512 hidden units at each layer and a 256-dimensional output vector.  The model is trained in a completely end-to-end fashion with both types of losses being back-propagated simultaneously.  Specifically, for any single example, we will compute a \emph{single class-conditional} NVP loss corresponding to the ground truth label for the sample, as well as the discriminative loss.  The losses are equally weighted, summed, and used in a standard backpropagation step.  Note that the discriminative loss is also propagated through the NVP layers, forcing the NVP layers to both separate the classes and model their densities correctly. 

Comparative ROC curves for our baseline class-conditional model are shown in Fig. \ref{fig:branched_roc}; the ROC for all data is shown in a solid line. We select ``uncertain'' samples on the basis of the size of the 95\% credible set inferred from the posterior distribution estimated using point density estimates as a proxy for counts in a beta-binomial model.  Where the range of that 95\% credible set exceeds 0.1,  we remove the sample from the test set as being `too uncertain' for the predictions to be reliable.  For example, a 95\% credible set of 0.33-0.36 (range of 0.03) would be retained, while a credible set of 0.13-0.55 (range of 0.42) would be removed. This resulted in approximately 120,000 points being rejected from the test set as `too uncertain to classify'. The remaining points are then re-plotted in an ROC plot, shown as a dashed line.  Note that the same samples -- those selected on the basis of the credible set size from the class-conditional model -- are removed from each plot. Removing test points based on the credible intervals derived from the CCCP-DE improves the ROC for the remaining test data when it is evaluated with either the class-conditional head based classification (done via ratio test), the class-conditional sigmoid head, or the standard FFNN.  The fact that removing points that CCCP-DE identifies as uncertain leads to improvements in both a CCCP-DE and FFNN model suggests that the uncertainty estimates are `consistent' in the sense that they are not dependent on the particular model with which they are evaluated and that we are in fact identifying points with intrinsically high uncertainty.

\section{Discussion}
\label{sec:discussion}

In Sec. \ref{sec:experiments}, we provided experimental evidence of our approach's capability to quantify uncertainty in a binary detection context. A natural extension to the multiclass regime of $M$ classes would leverage an $M$-headed NVP model, optionally in conjunction with a softmax classifier arm, with outputs from the $M$-headed model fed into a Dirichlet-Multinomial conjugate prior model. However, deriving credible sets is more challenging. While credible sets are $1$-dimensional regions for binary outputs, for a more general $M$-headed multinomial model, the credible interval becomes an $M-1$-dimensional region. While we could compute marginal credible sets on a per-class basis, this ignores potential covariance in errors, and leads to an uncertainty model that is difficult to interpret. A better way to leverage this conjugate prior model is an important topic for future research. 

A pressing concern, raised by \cite{nalisnick2018deep} is a flow-based model's ability to discriminate between samples from the training class distribution (in-set) and samples outside of the training class distribution (out-of-set). To examine this, using a similar model topology to Sec. \ref{sec:experiments} -- with a different input size and number of output heads -- we first fit a model to the Extended MNIST data set \cite{cohen2017emnist}, where we trained class-conditional density models on each of the 10 numeric digits 0-9 within the training split of the data, and used all characters of the test split (numeric and alphabetic) as a mixture of in-set and out-of-set samples for testing. For comparison with our earlier experiments on the PE dataset, we do not use a convolutional structure, which would likely improve the presented results for the EMNIST dataset. We also noted occasional instability in training, which no combination of regularization or model structure alteration appeared to adequately resolve.

Generative samples from the model are shown in Fig. \ref{fig:EMNIST-generative-samples} of the Appendix. We sampled from an isotropic multivariate Gaussian at each of the ten class-conditional heads, passed the result back through the network, then reshaping the image to the original 28x28 dimensions.  Despite the lack of a convolutional structure, the model has nevertheless learned a distribution over a reasonably diverse set of the digits.

We then evaluated the maximum log-likelihood across all heads for the test split.  In-set samples were expected to have a higher log-likelihood than out-of-set samples; while this was generally true (see Fig. \ref{fig:NLL_comparison} of the Appendix) 
we noted that there were many out-of-class points that scored relatively highly. This is reflected in a 0.7327 area under the curve (AUC) of a receiver operating characeristic (ROC) curve (cf. Fig. \ref{fig:ROC} of the Appendix) in which we attempted to distinguish in-set from out-of-set examples based on the likelihood -- while adequate for such a difficult task, there is still significant room for future performance improvements.



\begin{figure}[t!]
\centering
\includegraphics[width=\linewidth]{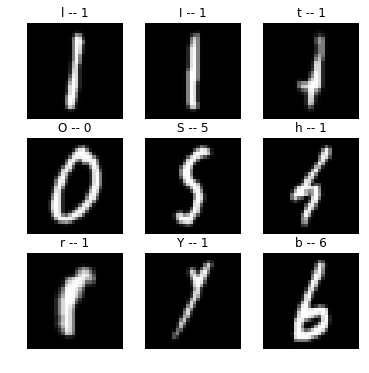}
\caption{A selection of out-of-set samples mistakenly identified as in-set; examples are labeled above with the actual class and predicted class on the basis of log-likelihood. For clarity: from left to right, top to bottom, ground truth classes are l (lower case ``L''), I (upper case ``i''), t, O, S, h, r, Y, and b.}
\label{fig:errors}
\end{figure}

When we examined the errors produced by using log-likelihood as an in-class/out-of-class signal, however, we found that the majority of the errors are from relatively easily confused classes.  To examine the errors more closely, we set a threshold to achieve 25\% False Positive Rate (labeling a sample as in-class when it was out-of-class) and selected the errors from approximately 110,000 test samples.  We found that 17\% of the errors come from the letter `l' (lower-case L) being mistakenly identified as in-class, most frequently identified on the basis of both likelihood and the discriminative portion as a numeric `1'.  The next most common was the character `I' (capital i) misidentified as being in-class (12\% of all false positives).  Other common false positives were the letter `t' (mistaken for 1 or 7), `O' (mistaken for numeric 0), `S' mistaken for `5', each making up approximately 9\% of errors.   A selection of misidentified samples is provided in Fig.~\ref{fig:errors}. These findings suggest a practical necessity to delineate between \textit{clear errors} and \textit{plausible mistakes} when assessing the quality of density estimators for in-class vs. out-of-class determination. 

\section{Conclusion}

In this paper, we have categorized uncertainty into multiple types. Most methods for dealing with uncertainty do not delineate different uncertainty types and address only one or two types of uncertainty. We are the first, to our knowledge, to introduce a model designed from first principles to jointly addresses all three of the aforementioned uncertainty types.  

While contemporaneous work  \cite{nalisnick2018deep} has raised concern about the ability of flow-based models to adequately model the density of data, particuarly with respect to outliers, we note that they attempted to model multi-modal data (all classes) with a single latent Gaussian distribution. Our empirical experience leads us to suspect that this is in fact not an appropriate strategy, as the fact that the model is transforming distinct clusters (in the input space) into a single isotropic Gaussian distribution, means that inevitably some region between the clusters (in the original feature space) must have very high density as well; see, for example, the lines of high density connecting the clusters in Fig. \ref{fig:rnvp}.  Our approach, in which each class is assigned to a single class-specific Gaussian output and these classes are driven to be separated in intermediate spaces by the cross-entropy classification head, mitigates this concern somewhat, although it does not account for potential multi-modality within a single class. We defer a detailed examination of this phenomenon to future work. We also note that the distinction between \textit{clear errors} and \textit{plausible mistakes} should be made to accurately assess the utility of a generative model for in-class vs. out-of-class determination.

Our models are efficient to train end-to-end and require only one forward pass per-sample to yield both a prediction and an uncertainty estimate. For the binary detection model, priors and base rates are also trivially incorporated for the binary detection problem and our stacked model provably bounds open space risk \cite{scheirer2013toward}  assuming that the density in the latent space converges to a Gaussian. The design naturally extends to a multiclass problem under a multinomial noise model with a Dirichlet conjugate prior, though for reasons discussed in Sec. \ref{sec:discussion} formalizing uncertainty estimates in this regime is a topic of future work.

The use of neural networks as generic parameter estimators is not new. However, estimating beyond the mean of a distribution is not extremely common in modern machine learning literature, leading us to surmise that there are many opportunities to take the concepts  presented herein to address a number of modern ML problems. Generalized linear models have been applied by statisticians for years, but they have not been widely adapted into deep neural network literature. In a sense, we can think of our approaches as applying \textit{really generalized linear models} using deep neural networks in place of link functions to allow for training and fitting.

While we have discussed in detail how to obtain uncertainty estimates, a related topic is what to do with uncertainty estimates once we have them. Active learning and semi-supervised learning are immediate applications, whereby the model's uncertainty estimates can be used to prioritize how to label samples on a budget or which types of samples to label on a budget. Additionally, inspecting samples that are misclassified and ascribed low uncertainty by the model in a validation set may shed interesting light on modes of failure in the model. Examining uncertainties associated with fooling or adversarial inputs is another direction for future research.
\section*{Acknowledgement}
This research was sponsored by Sophos PLC.

\bibliographystyle{icml2019}
\bibliography{references}

\newpage
\onecolumn
\twocolumn
\counterwithin{figure}{section}
\appendix
\section{Appendix}

\begin{figure}[h!]
\centering
\includegraphics[width=\linewidth,height=0.25\textheight,keepaspectratio]{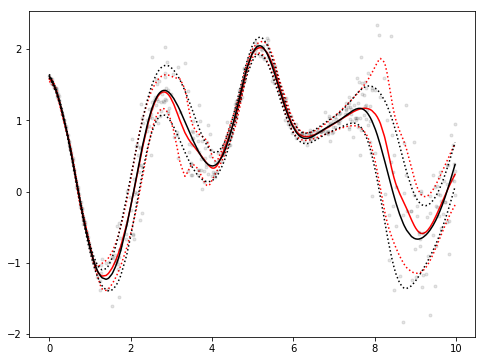}
\caption{Points generated from a Gaussian process with true mean shown in solid black and standard deviation shown as dashed black lines. Red solid and dashed lines represent the mean and standard deviation estimates from a neural network trained using a Gaussian negative log likelihood loss. While the manner by which uncertainty is estimated is not statistically principled, results are qualitatively correct, and can be obtained for a wide range of hyperparameter settings.}
\label{fig:gaussian_glm}
\end{figure}

\begin{figure}[h!]
\centering
\includegraphics[width=\linewidth,height=0.3\textheight,keepaspectratio]{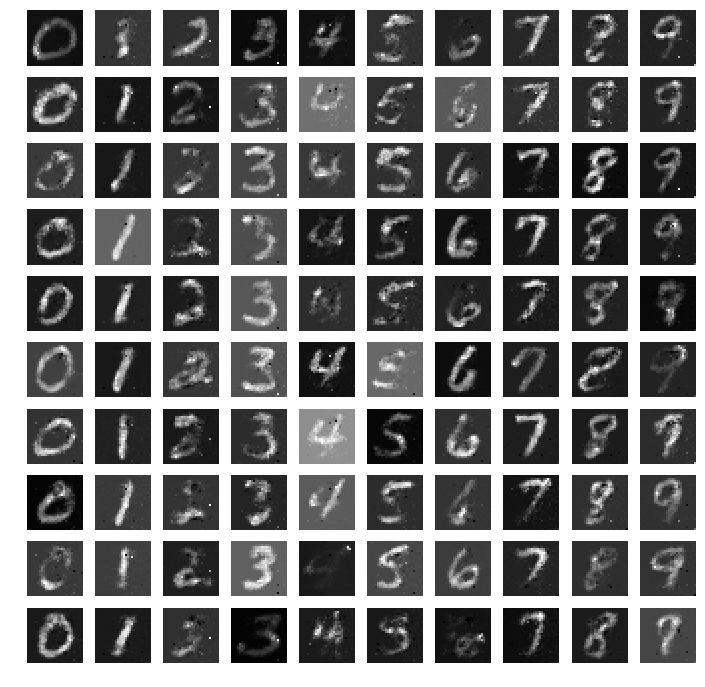}
\caption{Ten samples from each of the ten classes the CCCP-DE model was trained on.}
\label{fig:EMNIST-generative-samples}
\end{figure}

\begin{figure}[h!]
\centering
\includegraphics[width=\linewidth,height=0.3\textheight,keepaspectratio]{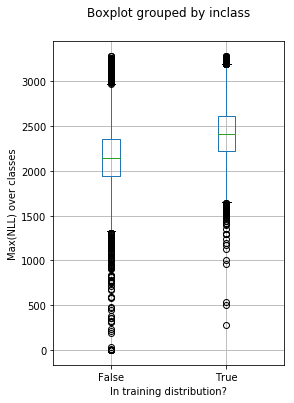}
\caption{In-set (digits) versus out-of-set (alphabetic) character negative log-likelihoods.  Note that these were evaluated across the test split of the EMNIST data, separate from the training split that was used to train the model.}
\label{fig:NLL_comparison}
\end{figure}

\begin{figure}[h!]
\centering
\includegraphics[width=\linewidth,height=0.3\textheight,keepaspectratio]{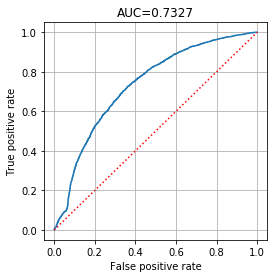}
\caption{ROC plot for the binary classification task of distinguishing in-set from out-of-set examples on the basis of sample log-likelihood.}
\label{fig:ROC}
\end{figure}

\end{document}